\pdfoutput=1
\PassOptionsToPackage{table}{xcolor}
\documentclass[11pt]{article}

\usepackage{acl}

\usepackage{times}
\usepackage{latexsym}

\usepackage[T1]{fontenc}
\usepackage[utf8]{inputenc}

\usepackage{microtype}

\usepackage{inconsolata}

\usepackage{graphicx}
\title{Improving the Inclusivity of Dutch Speech Recognition by Fine-tuning Whisper on the JASMIN-CGN Corpus}

\author{Golshid Shekoufandeh \\
  University of Amsterdam, \\
  the Netherlands \\
  \texttt{g.shekoufandeh@uva.nl} \\\And
  Paul Boersma \\
  University of Amsterdam, \\
  the Netherlands \\
  \texttt{paul.boersma@uva.nl} \\\And
  Antal van den Bosch \\
  Utrecht University, \\
  the Netherlands \\
  \texttt{a.p.j.vandenbosch@uu.nl} \\}
  
\begin{document}
\maketitle
\begin{abstract}
We test and study the variation in speech recognition of fine-tuned versions of the Whisper model on child, elderly and non-native Dutch speech from the JASMIN-CGN corpus. Our primary goal is to evaluate how speakers' age and linguistic background influence Whisper's performance.
Whisper achieves varying Word Error Rates (WER) when fine-tuned on subpopulations of specific ages and linguistic backgrounds. Fine-tuned performance is remarkably better than zero-shot performance, achieving a relative reduction in WER of 81\% for native children, 72\% for non-native children, 67\% for non-native adults, and 65\% for native elderly people.
Our findings underscore the importance of training speech recognition models like Whisper on underrepresented subpopulations such as children, the elderly, and non-native speakers.\\
\end{abstract}

\section{Introduction}
This study explores the challenges of Automatic Speech Recognition (ASR) in accurately transcribing speech from underrepresented subpopulations in the Netherlands, specifically non-native, child and elderly speech. Older ASR systems were often trained on datasets mainly consisting of adult speakers with standard pronunciation, and therefore struggled to accommodate the acoustic and linguistic characteristics of more diverse subpopulations. Recent advances in ASR have shifted away from this ``one-size-fits-all'' approach and aim to build models capable of recognizing and processing speech with greater inclusivity.

Recent studies have highlighted that state-of-the-art ASR systems exhibit varying performance between different groups of speakers, influenced by factors such as gender, age, and accent.
The efficacy of ASR systems is influenced by many properties of speakers, such as their age and whether they are native speakers of the language in question \citep{feng2024towards}.  
Previous studies have shown that training or fine-tuning ASR models on subsets of data specific to these subpopulations can yield performance improvements. For example, research on regionally-accented and foreign-accented English speech (end-to-end training with GRUs: \citealt{viglino2019end}; the challenge organized by \citealt{shi2021accented}; further \citealt{sullivan2022improving} and \citealt{shibano2021speech}) show that such adaptations tend to enhance ASR accuracy. After these findings for English diverse speech, our project is the first to fine-tune a transformer model on diverse Dutch speech. 

Given the rapidly evolving technological landscape, accessibility of users to ASR systems is of paramount importance.  Existing ASR systems often struggle to capture the real-world features of speech, as the Whisper model has used, for example, Common Voice and LibriSpeech, which consist entirely of read speech. Real-life conversations differ from this in many respects, and we are fortunate in having at our disposal a database that contains more natural (interactive, spontaneous) modes of speech, namely the JASMIN-CGN corpus.

Recognizing the substantial variability in speech patterns and linguistic characteristics due to factors such as age and native language, this study hopes to contribute to the development of ASR systems that recognize diverse patterns of speech, including both native and non-native speakers and various ages.

\section{Description of the Dataset}
For this study, we used the JASMIN-CGN corpus \citep{cucchiarini2006jasmin}. This corpus is an extension of the CGN, or Spoken Dutch Corpus~\citep{oostdijk2000spoken}, broadening its scope in terms of age, mother tongue, and communication setting. The JASMIN-CGN corpus encompasses a wide range of demographic groups, capturing Dutch speech data from children of varying ages, elderly individuals, and non-native speakers.

One aspect of this corpus is its inclusion of human-machine interaction scenarios, which were not part of the original CGN. The corpus is divided into five speaker groups: native children in primary and secondary school, non-native children and adults, and senior citizens. Each group provides two types of speech data: read and interaction-based, contributing to a total of approximately 90 hours of Dutch data, of which 60 hours come from the Netherlands and 30 hours from Belgium. We restrict ourselves to the Netherlands here. The demographic breakdown and the corresponding data hours for each group are detailed in Table \ref{tab:jasmin}.
\begin{table*}
    \footnotesize
    \centering
    \begin{tabular}{cccc} 
         Group&  demographic categories&  Age range&  Hours of speech\\ 
         1&  Native Young Children&  7 to 11 years&  12h 21m\\ 
         2&  Native Teenagers&  12 to 16 years&  12h 21m\\ 
         3&  Non-native Children&  7 to 14 years&  12h 21m\\ 
         4&  Non-native Adults (L2 learners)&  18 to 60 years&  12h 21m\\ 
 5& Native Elderly& 65 years and above& 9h 26m\\ 
    \end{tabular}
    \caption{\textit{Demographic categories and characteristics of speech samples in JASMIN-CGN from the Netherlands. This table presents demographic groupings along with the corresponding age ranges and total hours of speech data collected for each subpopulation.}}
    \label{tab:jasmin}
\end{table*}

\section{Data Preparation}

\subsection{Segmentation}
The preparation of the data set for training involved converting transcriptions stored in Latin-1-encoded .ort files into UTF-8-encoded plain text files for compatibility with the selected ASR model. For audio recordings exceeding 30 seconds, a meticulous segmentation process was automated using Python scripts that aligned the audio segments with their respective transcriptions using the timestamps provided in the .ort files.

\subsection{Data Cleaning}
The documentation of the JASMIN dataset \citep{cucchiarini2006jasmin} mentions that in case of mispronunciations, foreign language terms or dialect usage by speakers, they have used specific codes prefixed with ``*''. We have removed these codes and their prefix, as Whisper can transcribe them. Some examples of these codes include the following:
\begin{itemize}
    \item *v: Denoting words from a foreign language.
    \item *n: Indicating new words or interjections.
    \item *a: Representing words cut short or interrupted.
\end{itemize}

\section{Methodology}
One of the essentials of speech recognition is speaker diversity. The study by \citet{wilpon1996study} underscores the importance of diversity, revealing that automatic speech recognition technology at the time did not perform well for children and elderly people.  This required systems developed specifically for these demographics to ensure inclusivity and effectiveness. \citet{sullivan2022improving} provide evidence that training under a rich set of L1 and L2 conditions can improve recognition accuracy for non-native speakers. \citet{shibano2021speech} emphasize the importance of inclusivity in speech recognition, arguing that systems should be designed to work well for all users, not just a subset. These papers collectively illustrate the necessity and effectiveness of training speech recognition systems on data from underrepresented groups such as children, the elderly, and non-native speakers. In this section we delve into the methodology employed for the fine-tuning of the Whisper model. We use this process to enhance its performance on speech from children, the elderly, and non-native speakers.

\subsection{Whisper Model}
OpenAI's Whisper large-v2 model, a sequence-to-sequence model based on the Transformer architecture \citep{NIPS2017_3f5ee243}, is designed for automatic speech recognition and speech translation \citep{whisper2023}. It was trained using a 680,000-hour data set of labeled speech data, using large-scale weak supervision techniques \citep{whisper2023}.

The architecture of Whisper includes stacked encoder and decoder blocks and an attention mechanism, characteristic of the Transformer model. It handles audio recordings in segments of 30 seconds. Like many speech recognition systems, Whisper begins by converting raw audio into a more digestible format. This is achieved by transforming the audio signal into a Mel spectrogram, which is a visual representation of how frequencies change over time. The Mel spectrogram can be thought of as an ``image'' of the sound, represented by a matrix of numbers. When visualized, it provides a continuous representation of auditory frequency over time, making it visually interpretable.

The Whisper large-v2 model can be accessed on Hugging Face\footnote{\url{https://huggingface.co/openai/whisper-large-v2}}, and its corresponding source code is available on GitHub\footnote{\url{https://github.com/huggingface/transformers/issues/20653}}.

Our research primarily utilizes the Whisper model, drawn to its capabilities and features that proved to be particularly advantageous for our study involving Dutch datasets. The Whisper model stands out for its robustness in handling diverse and complex linguistic structures, a feature that is crucial when dealing with Dutch, a language known for its intricate syntax \citep{dutchmorphology}. Furthermore, Whisper’s advanced noise reduction capabilities ensure high-quality transcription even under less-than-ideal acoustic conditions, a common scenario in real-world data collection. Although other open-source alternatives like Dutch ASR \citep{povey2011kaldi} and wav2vec \citep{baevski2020wav2vec, conneau2020unsupervised, babu2021xls} have their own merits, we had two specific reasons for not using them for fine-tuning.

First, our goal was to perform an exhaustive cross-validation analysis for each group. This required the development of numerous models, and incorporating fine-tuning across multiple systems would have significantly amplified the study's workload and complexity.

Second, most existing Dutch ASR systems are built on Kaldi, a widely used ASR toolkit known for its flexibility and extensibility. However, the field of speech recognition is rapidly advancing with the emergence of new methods and technologies. One reason for our shift away is the time-intensive nature of training Kaldi. In contrast, the Whisper model, having been pre-trained on extensive datasets, requires fewer data for fine-tuning, making it ideal for tasks with limited datasets.

\subsection{\texorpdfstring{$k$}{Lg}-Fold Cross-Validation}
\label{kfold}
A methodology of $k$-fold cross-validation was implemented, with $k = 10$. This entailed the creation of 10 distinct folds (for each of the four datasets), each serving as a unique combination of a test set (10\% of the data, different for each fold) and a combined training and validation set (the remaining 90\% of the data, largely overlapping between folds). The motivations behind employing $k$-fold cross-validation were (1) to provide a somewhat accurate estimate of performance by averaging over the 10 folds of the relatively small dataset, and (2) to obtain an estimate of the variance of the system's performance over different test sets.

\subsection{Partitioning the Training and Validation Set}
After forming the combined training and validation set (90\% of the data in a fold), the subsequent step involved dividing this set into distinct training and validation subsets. The division was structured in such a way that the training set comprised 80\% of the data, while the validation set constituted 10\%.

The validation set helped us to monitor the model's performance level during the training process. We created checkpoints (snapshots) of the model after approximately every 0.1 epoch. We used the validation set to compute a WER for every checkpoint, after which we selected the checkpoint with the lowest WER to be our ``best'' model for the fold at hand. We then put this ``best'' model to the definitive test; that is, we measured the WER that our selected ``best'' model yielded on the test set, thus giving a definitive estimate of how well our training procedure for this fold generalized to genuinely unseen data.

\subsection{The Five Experiments}
The whole study involved five fine-tuning experiments, each starting with the original Whisper large-v2 model.

Each of the first four experiments was dedicated to one of the demographic groups. Each experiment on a demographic group involved training the Whisper model 10 times using a fold of data of that group. We used a constant learning rate of \(3\cdot10^{-5}\). For each fold, we trained the model for five epochs through the training data, a duration that the authors of Whisper considered sufficient for optimal performance \cite{whisper2023}.

The fifth fine-tuning experiment was applied to the combined dataset.

\subsection{Evaluation}
The fine-tuned models were evaluated based on the Word Error Rate (WER), a commonly used metric that considers the number of substitutions, deletions, and additions between two transcripts. The performance on a test set is recorded using the fine-tuned model checkpoint that yields the lowest WER on the validation set.
\section{Results}
\begin{table*}[htbp]
    \scriptsize
    \centering
    \resizebox{\textwidth}{!}{%
    \begin{tabular}{lcccc}
              &  Tested on Native children&  Tested on Non-native children&  Tested on Non-native adults&  Tested on Native elderly\\
              Zero-shot evaluation (no fine-tuning)&  26.12\%&  38.48\%&  42.07\%&  28.73\%\\
              Fine-tuned on native children& {\cellcolor{lightgray!60}}$5.45 (\pm 0.50)\%$&  $26.69(\pm0.15)\%$&  $28.37(\pm 0.42)\%$&  $23.19(\pm 0.24)\%$\\
              Fine-tuned on non-native children&  $21.77(\pm 0.22)\%$& {\cellcolor{lightgray!60}}$13.13(\pm 1.11)\%$&  $26.14(\pm 0.31)\%$&  $24.00(\pm 0.55)\%$\\
              Fine-tuned on non-native adults&  $24.27(\pm 0.41)\%$&  $27.57(\pm 0.11)\%$&  {\cellcolor{lightgray!60}}$14.01(\pm 2.23)\%$&  $24.14(\pm 0.63)\%$\\
              Fine-tuned on native elderly&  $23.09(\pm 1.15)\%$&  $29.24(\pm 0.09)\%$&  $28.19 (\pm 0.60)\%$&  {\cellcolor{lightgray!60}}$\mathbf{9.11(\pm 0.73)\%}$\\
      Fine-tuned on all of the data& {\cellcolor{lightgray!60}}$\mathbf{4.98(\pm 0.42)\%}$& {\cellcolor{lightgray!60}}$\mathbf{10.90(\pm 0.88)\%}$&{\cellcolor{lightgray!60}} $\mathbf{13.95(\pm 3.90)\%}$&{\cellcolor{lightgray!60}}$9.96(\pm 1.91)\%$\\
    \end{tabular}
    }
    \caption{\textit{Mean and standard deviation Word Error Rate (WER) of the folds for fine-tuned Whisper model across diverse groups.}}
    \label{tab:table}
\end{table*}

A first evaluation was conducted on the zero-shot Whisper model for four subsets. The results of this evaluation can be found in the first row of Table \ref{tab:table}.

\subsection{Fine-tuning and Testing on the Four Distinct Subsets}
After the zero-shot evaluation, the model was fine-tuned using 10-fold cross-validation (§\ref{kfold}) separately on four distinct subsets: native children, non-native children, non-native adults, and native elderly individuals. Each grey cell on the diagonal of Table \ref{tab:table} shows the WER performance for one finetuned subset, averaged over the 10 folds (each fold comes with a ``best'' model, selected with the help of the validation set and tested on the appropriate held-out test set). Relative improvements in WER are by 81\% for native children (i.e. from 26.12\% to 5.45\%), 72\% for non-native children, 67\% for non-native adults, and 65\% for native elderly people (the odds ratios of the improvements are 6.1, 4.1, 4.5, and 4.0, respectively). Thus, fine-tuning gave the four groups large WER improvements, with the largest improvement for native children (from 26.12\% to 5.45\%). 

\subsection{Fine-tuning and Testing on the Full Dataset}
In the final (fifth) experiment, 10 folds were used to create a unified dataset to fine-tune the Whisper model. However, this approach introduced certain limitations, including a bias towards the native children's group due to its larger size. A more effective approach could have involved merging each 10-fold of one group with each 10-fold of the other groups (= 100 fine-tuning experiments), but time constraints precluded this. 

After fine-tuning, tests were conducted on the respective test sets associated with each group's fold.\footnote{For example when we combine the training data from fold 4 of each group, we will evaluate the fine-tuned model using the test set that corresponds to each group’s fold 4.} The results of these evaluations are detailed in the bottom row of Table \ref{tab:table}.\footnote{Fine-tuning the Whisper model poses some challenges, including occasional hallucinations that hinder a comprehensive evaluation of its potential. Specifically, the model tended to hallucinate the Unicode replacement character (U+FFFD), prompting a re-evaluation of the data transcription encoding. This issue has been resolved by mentioning the language of transcription in the whisper evaluation pipeline.} We see that these results are on average at least as good as those on the diagonal, from which we conclude that training all four groups together does not yield worse results for a single group than training that single group alone (if anything, the result may be a bit better).

\subsection{Transfer of Performance to Other Subsets}
Looking at what happens off the diagonal in Table \ref{tab:table}, we see that fine-tuning on one group tends to improve the recognition of all other groups. This can be a corpus effect: as JASMIN-CGN contains interactive and spontaneous speech, with many short sentences, simply training on such sentences may help the recognition of other such sentences, even when spoken by people with different ages and/or degress of nativeness than the fine-tuned group.

Looking more closely, we see especially that the recognition of non-native child and non-native adult speech improves appreciably even by fine-tuning a model on native children or native elderly: the performance on non-native children improves from 38.48\% to 26.69\% or 29.24\%, and the performance on non-native adults improves from 42.07\% to 28.37\% or 28.19\%. This could partly be due to the fact that the zero-shot performance on the non-native groups was relatively poor (hence, more room for improvement), and partly due to the possibility that age variation improves generalizability more than nativeness variation does (because fine-tuning on non-native adults or children doesn’t improve the recognition of native children or elderly much).

\subsection{Bottom Line}
There are some indications that fine-tuning on various ages helps a bit more than fine-tuning on various degrees of nativeness, but it is clear that the best results are achieved by training on a comprehensive dataset that includes both age variation and nativeness variation.

\section{Discussion}
This study encompassed an evaluation of Whisper
models that were fine-tuned across different demographic groups, revealing that:
(1) fine-tuning on a group helps the recognition of that group tremendously;
(2) fine-tuning on a group also moderately helps the recognition of all other groups (especially the non-natives);
(3) fine-tuning on all groups together helps the recognition of each group at least as much as fine-tuning on that group alone.
So we recommend fine-tuning on as many diverse subpopulations as possible.

Using Whisper models, substantial improvements in speech recognition accuracy were observed, particularly among native speakers. This underscores the potential to take advantage of each speaker's characteristics to enhance the overall performance of ASR systems. Furthermore, the study highlights the challenges associated with demographic-specific fine-tuning in ASR systems, laying the groundwork for future research endeavors.

\section{Conclusion}
In this research, using the JASMIN-CGN corpus, we explored how a large pre-trained ASR model can adapt to variations in age and language proficiency, for Dutch as spoken in the Netherlands. By comparing subpopulations of different ages and degrees of nativeness, we came to understand how these variables influence the final model's ability to accurately transcribe speech.
The goal of this study was to identify the factors that affect speech recognition accuracy in different subpopulations, which will help us create more inclusive systems.
Furthermore, we showed that compared to the read sentences of Common Voice and LibriSpeech, on which Whisper was pre-trained, there exist more natural, speech-like resources, such as CGN and JASMIN-CGN, which can help fine-tune ASR models to better adapt to real-life settings that involve spontaneous interactions among humans or between humans and machines.

%\section*{Acknowledgements}

\bibliography{main}

%\appendix

%\section{Example Appendix}
%\label{sec:appendix}

\end{document}